\documentclass{article}

\usepackage{arxiv}

\usepackage[utf8]{inputenc} 
\usepackage[T1]{fontenc}    
\usepackage[colorlinks=true,allcolors=black]{hyperref}       
\usepackage{url}            
\usepackage{booktabs}       
\usepackage{amsfonts}       
\usepackage{nicefrac}       
\usepackage{microtype}      
\usepackage{lipsum}		
\usepackage{graphicx}
\usepackage[numbers]{natbib}
\usepackage{doi}
\usepackage{pifont}
\usepackage{xcolor}
\usepackage{amsmath}
\usepackage{fancyhdr}
\usepackage{array}
\usepackage{multirow}
\usepackage{colortbl}

\definecolor{dpo-color}{HTML}{44AA99}
\definecolor{gpt-4o-color}{HTML}{DDCC77}
\definecolor{distill-color}{HTML}{332288}

\title{Training LLM-based Tutors to Improve Student Learning Outcomes in Dialogues}

\author{Alexander Scarlatos\\
University of Massachusetts Amherst\\
ajscarlatos@cs.umass.edu\\
\And
Naiming Liu\\
Rice University\\
nl35@rice.edu\\
\And
Jaewook Lee\\
University of Massachusetts Amherst\\
jaewooklee@cs.umass.edu\\
\And
Richard Baraniuk\\
Rice University\\
richb@rice.edu\\
\And
Andrew Lan\\
University of Massachusetts Amherst\\
andrewlan@cs.umass.edu
}

\date{}

\renewcommand{\headeright}{}
\renewcommand{\undertitle}{Published in AIED 2025: The 26th International Conference on Artificial Intelligence in Education}

\begin{document}
\maketitle

\begin{abstract}
Generative artificial intelligence (AI) has the potential to scale up personalized tutoring through large language models (LLMs). Recent AI tutors are adapted for the tutoring task by training or prompting LLMs to follow effective pedagogical principles, though they are not trained to maximize student learning throughout the course of a dialogue. Therefore, they may engage with students in a suboptimal way. We address this limitation by introducing an approach to train LLMs to generate tutor utterances that maximize the likelihood of student correctness, while still encouraging the model to follow good pedagogical practice. Specifically, we generate a set of candidate tutor utterances and score them using (1) an LLM-based student model to predict the chance of correct student responses and (2) a pedagogical rubric evaluated by GPT-4o. We then use the resulting data to train an open-source LLM, Llama 3.1 8B, using direct preference optimization. We show that tutor utterances generated by our model lead to significantly higher chances of correct student responses while maintaining the pedagogical quality of GPT-4o. We also conduct qualitative analyses and a human evaluation to demonstrate that our model generates high quality tutor utterances.\footnote{Our code is available at \url{https://github.com/umass-ml4ed/tutorbot-dpo}}
\end{abstract}


\section{Introduction}

Recent advances in generative artificial intelligence (AI), including large language models (LLMs), have opened new possibilities in education and in particular on scaling up personalization. One form of personalization that generative AI powers is interactive learning via \emph{tutoring dialogues} between AI-powered tutors and students. These interactions have the potential to tailor instruction to each student's needs and progress, while offering personalized feedback, all in real time, in a scalable way. 
Given the widespread success of human tutors for improving student outcomes~\cite{metaat}, many recent works have developed LLM-based tutors, showing promise across various educational domains~\cite{han2024llm,lieb2024student,nye2023generative,pal2024autotutor,park2024empowering,schmucker2024ruffle,stamper2024enhancing,yang2024cref}. Many LLM-based tutors are even deployed in practice, such as Khan Academy’s Khanmigo \cite{khanmigo} and Carnegie Learning’s LiveHint \cite{livehint}. 
Several preliminary studies have shown that interacting with LLMs can increase student learning~\cite{zhang2024future}, although some have shown that students can develop an over-reliance on LLMs which negatively impacts their learning~\cite{krupp2024challenges}.

Many prior works have focused on improving LLMs' ability to follow effective tutoring principles, adapting them for the tutoring task that they are not pre-trained for. One approach, explored in \cite{wang2024bridging}, analyzes the decision-making process underlying human tutor utterances, showing that integrating expert decisions enhances LLM-based tutoring. Another study, \cite{macina-etal-2023-mathdial}, examines tutor moves in interactions with an LLM-powered simulated student agent, demonstrating that move annotation data contributes to better tutoring performance. Similarly, \cite{learnlm} investigates the role of AI roleplay in generating synthetic tutoring data and finds that fine-tuning LLMs on this data, along with human tutor-student interactions, significantly improves their pedagogical effectiveness. Moreover, \cite{demszky2021measuring} introduces the concept of tutor uptake---acknowledging student responses---as a valuable strategy for LLMs to adopt. 

While these efforts offer valuable insights into how LLMs can emulate effective human tutoring strategies, the question remains whether such approaches truly maximize student learning outcomes. 
Rather, a data-driven approach, where student outcomes form a reward signal, could potentially lead to AI tutors that are more aligned with educational goals. Additionally, current approaches often rely on large proprietary LLMs, coming with many downsides; they are not fully controllable and cannot be easily customized, can be costly to use, and relinquish control of private student data. 
Therefore, developing effective AI tutors with smaller, open-source LLMs remains an important goal.

\subsection{Contributions}
In this paper, we propose a novel approach to train a small, open-source LLM to not only follow effective pedagogical principles, but directly optimize for student learning outcomes. Our method performs three key steps. First, at each dialogue turn, we gather multiple candidate tutor utterances from a variety of sources, including human tutors and LLMs with varying styles and sizes. Second, we evaluate each candidate utterance on two aspects: (1) whether it elicits a correct response in the next student turn, using a trained student model for dialogues to predict student behavior, and (2) whether it adheres to a set of effective pedagogical principles, using GPT-4o in a LLM-as-a-judge evaluation setup. 
Third, we contrast good candidate utterances with poor ones, and fine-tune Llama 3.1 8B~\cite{dubey2024llama} with offline Reinforcement Learning (RL), specifically Direct Preference Optimization (DPO)~\cite{rafailov2024direct}. 
We demonstrate that our optimized LLM tutor significantly increases the likelihood of next turn student response correctness, while reaching comparable pedagogical quality to that of a much larger, proprietary LLM, GPT-4o~\cite{openai2024gpt4o}. Through qualitative analysis and human evaluation, we confirm that our approach produces high-quality tutor utterances and reveal emergent tutoring strategies that arise from our training approach. 

We acknowledge up front that the most significant limitation of our work is that we do not experiment with real students. Since access to students at the scale necessary for our work is beyond our capability, we use a simulated student model instead. While we believe that our work is a reasonable starting point to train LLM-based tutors to maximize student outcomes, future work with real students in the loop is highly important. Therefore, to facilitate further research, we publicly release our code and encourage researchers and practitioners with access to real-world tutoring dialogue settings to build on our work.

\section{Related Work}

\paragraph{AI Tutors in Dialogues}

There is a long history of AI-based tutors in education that interact with students through dialogues. Early systems, such as Cognitive Tutors, construct cognitive models of students to provide targeted feedback~\cite{anderson1995cognitive}. AutoTutor engages students by asking targeted questions, and assesses student correctness using latent semantic analysis~\cite{graesser2001intelligent}. Why2-Atlas converts a student response to a proof, which it uses to identify misconceptions and guide a dialogue~\cite{vanlehn2002architecture}. 
While these systems were often effective for improving student learning, they required significant engineering and had limited flexibility. In contrast, recent LLM tutors can more easily adapt to new contexts, interpret student responses, and cater personalized content towards the student. Several LLM tutors are implemented through refined prompt engineering~\cite{khanmigo}, with some taking on specialized roles such as teachable agents~\cite{schmucker2024ruffle} or ``co-pilots'' for human tutors~\cite{wang2024tutor}. Other works fine-tune LLM tutors to enhance the pedagogical capabilities over the base models. A common approach is to generate simulated dialogues, where the tutor utterances are constructed to follow good pedagogical practice, and fine-tune on those~\cite{sonkar-etal-2023-class,learnlm}. Several works also generate examples of low quality tutor utterances and use them as negative samples in DPO training to improve over fine-tuning~\cite{ashok-kumar-lan-2024-improving,sonkar-etal-2024-pedagogical}.

\paragraph{Student Outcome Modeling}

While there are many ways of measuring student outcomes, in this work we focus on the well-studied setting of \textit{next item correctness}. Student modeling in this setting is typically handled by knowledge tracing (KT), where a binary correctness is predicted for the next item a student attempts based on the student's history so far \cite{kt}. KT models have used recurrent neural networks \cite{dkt}, self-attention networks \cite{akt}, and, more recently, LLMs \cite{cui2023adaptive,okt}. A recent work introduces LLMKT \cite{scarlatos2024exploring}, an LLM-based model that predicts \textit{next turn} student correctness in dialogues. Therefore, we leverage LLMKT to predict student outcomes in this work. Similar to our work, prior works have used RL to discover teaching policies with rewards derived from student models \cite{he2021quizzing,rafferty2016faster}, including KT-based student models \cite{cai2019learning}. Another recent work used LLM estimates of student post-test scores to refine math worksheets \cite{heyueya2024evaluatingoptimizingeducationalcontent}. However, to the best of our knowledge, ours is the first to do so in the context of tutoring dialogues.

\paragraph{Evaluating Pedagogical Quality of LLMs}

To validate LLM tutors, we need to be able to evaluate them along pedagogical measures.
Typically, researchers construct a pedagogical \textit{rubric}, which defines multiple properties that tutor utterances should follow. Rubric-based evaluation of generated utterances is then performed by human experts \cite{jia2022insta,learnlm,wang2024bridging} or by LLMs \cite{kakarla2024using,scarlatos2024improving}. In this work, we design a pedagogical rubric and primarily use LLMs to evaluate tutor utterances, but also humans at a smaller scale to ensure our results are reliable. Similar to \cite{scarlatos2024improving}, we also use LLM-assigned rubric scores to form DPO preference pairs.

\section{Methodology}
\begin{figure}[t]
    \centering
    \includegraphics[width=.8\textwidth]{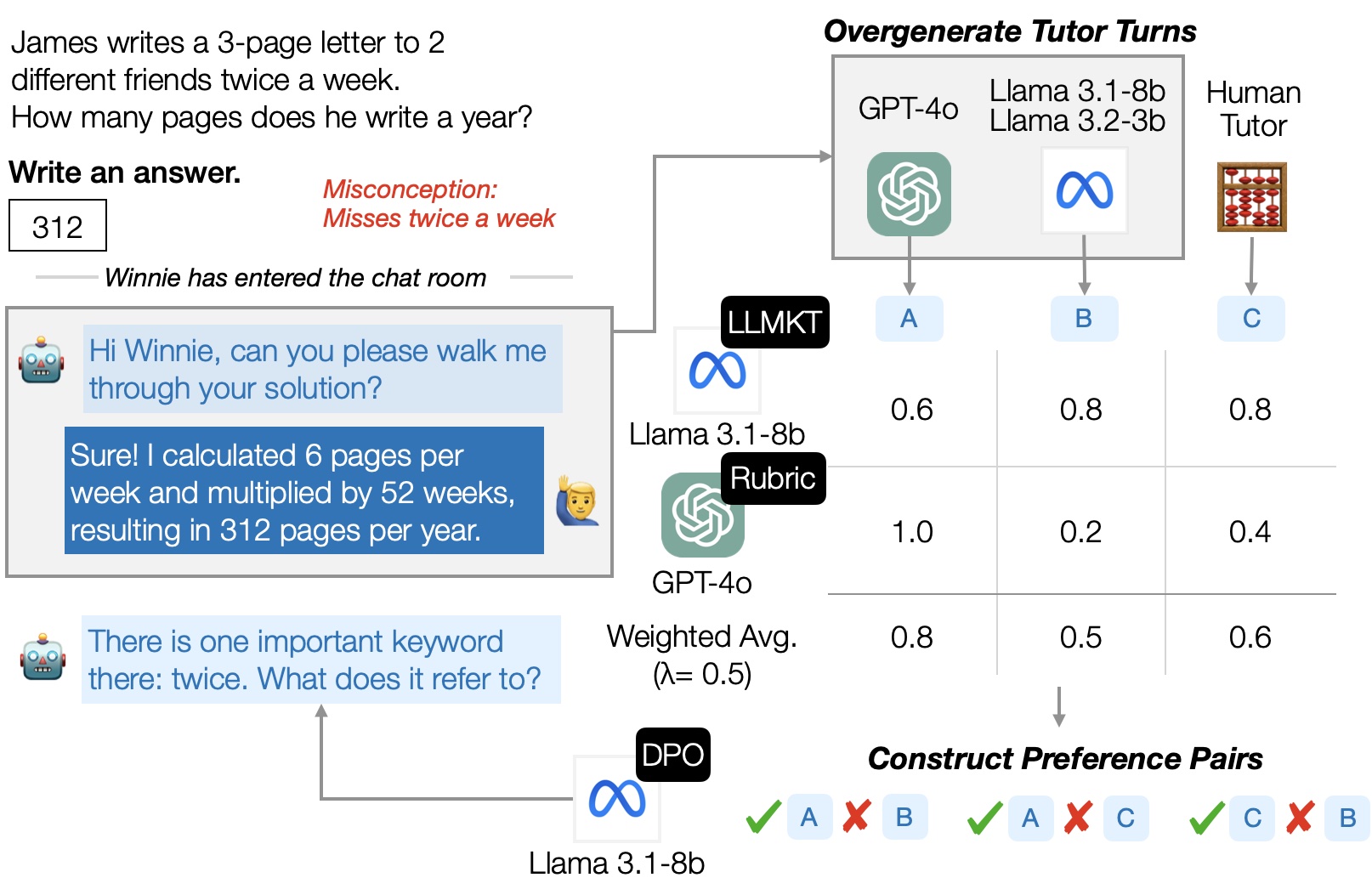} 
    \caption{An overview of our methodology for training LLMs to generate tutor utterances with the joint objective of maximizing student learning outcomes and following good pedagogical principles. \textit{Human tutor} refers to tutor utterances in MathDial.}
    \label{fig:overview}
\end{figure}

We now detail our methodology to generate tutor dialogue utterances to maximize student learning outcomes. Figure~\ref{fig:overview} shows an overview of our approach with an example, where the problem is from the MathDial~\cite{macina-etal-2023-mathdial} dataset.
In this scenario, a student has a misconception, missing the ``twice'' part of the problem, and provides an incorrect answer. 
We generate multiple, diverse candidate versions of the next tutor utterance to follow the student's response, using a collection of LLMs with different sizes and styles. 
Then, among the generated utterances and the human tutor utterance, we construct preference pairs to fine-tune an LLM using DPO. We consider an utterance to be preferred if it (1) likely elicits a correct student response and (2) follows good pedagogical principles. The former criterion employs a student model, LLMKT~\cite{scarlatos2024exploring}, to predict whether the student will correctly respond to the tutor in their next turn. The latter criterion employs a set of rubric items to evaluate whether a tutor utterance follows good pedagogical principles.
Before detailing each component, we define some key notations: a dialogue 
$d$ consists of an alternating sequence of tutor and student turns, $d=(t_1,s_1,\ldots,t_M,s_M)$, where $M$ is the number of turn pairs, indexed by $m$, and the textual content of each turn is the ``utterance''.

\subsection{Student Outcome Prediction}

We now detail the model we use for predicting student outcomes in dialogues.
We leverage recent work on knowledge tracing (KT) in dialogues, LLMKT~\cite{scarlatos2024exploring}, an LLM-based model that predicts if a student will respond correctly to a tutor-posed task in their next turn, given dialogue history and the knowledge components (KCs)~\cite{commoncore} embedded in a tutor turn. The model is highly accurate at predicting student correctness, achieving $0.76$ AUC on the MathDial test set, making it a reliable automated source for estimating student outcomes.

In LLMKT, at the $m$-th dialogue turn for the student, the model analyzes the conversation history $t_{\le m}, s_{<m}, \mathcal{C}_{m}$, which denote the tutor utterances up to and including this current turn, prior student turns, and the set of KCs involved in the tutor-posed task in the current turn, respectively. It estimates student knowledge levels on each KC, and combines these estimates to predict whether the student will respond to the tutor-posed task in the current turn correctly. Therefore, we leverage LLMKT as a \emph{student simulator} and use it to evaluate whether a generated tutor utterance at a dialogue turn will elicit a correct response from the student.

\subsection{Following Pedagogical Principles}
\label{sec:rubric}

In addition to promoting correct student responses, we also hand-craft a set of rubric items for effective pedagogical principles, listed Table~\ref{tab:ai_tutor_criteria}. We then employ GPT-4o to assess how well the generated tutor utterances align with the rubric. By incorporating pedagogical evaluation aspects rather than focusing solely on correctness prediction, our approach discourages oversimplification of tutor utterances and instead encourages utterances that provide meaningful guidance.

\begin{table}[t]
    \centering
    \caption{Pedagogical evaluation criteria for tutor utterances in dialogues.}
    \begin{tabular}{p{1.5cm}p{10cm}}
        \toprule
        \textbf{Criteria} & \textbf{Explanation} \\
        \midrule
        Accuracy & Ensuring the response does not contain false or misleading statements. \\
        \midrule
        Progress & Determining whether the response  helps the student move forward. \\
        \midrule
        \multirow{4}{*}{Guidance} & 1. Error identification: Correctly pinpoints the student's mistake. \\
        & 2. Strategic Hinting: New information or guidance for help. \\
        & 3. Withholding: Refrains from directly providing the final answer. \\
        & 4. Encouraging: Motivates the student to persist in their attempt. \\
        \bottomrule
    \end{tabular}
    \label{tab:ai_tutor_criteria}
\end{table}

Our evaluation draws inspiration from feedback assessment studies~\cite{jia2022insta,kakarla2024using,steiss2024comparing} and focuses on common errors made by LLMs when generating feedback for math problems~\cite{scarlatos2024improving}. The rubric evaluates generated tutor utterances on six granular items, each assigned a binary label, across three core aspects: \textit{Accuracy}, \textit{Progress}, and \textit{Guidance}.
Considering all aspects, we also have GPT-4o provide an \textit{overall score} for the utterance on a 1-10 scale. Our prompt leverages chain-of-thought so that GPT-4o provides reasoning about the utterance before assigning its scores.

We include the human tutor utterance from MathDial in our prompt as a point of comparison, mainly to assist with evaluating \textit{Accuracy}, which can be challenging \cite{scarlatos2024improving}. We find the 1-10 scale works slightly better than simply averaging all the binary rubric items, possibly because it enables GPT-4o to decide which rubric items are more relevant given the context of the dialogue.

\subsection{Preference Pair Construction}

To create a dataset of preference pairs for DPO training, we collect candidate tutor utterances from four sources: (1) the human tutor utterances from the MathDial dataset, (2) utterances generated by GPT-4o, where the evaluation criteria (rubric and intending to elicit correct student responses) are included in the prompt, (3) utterances generated by Llama 3.1 8B using a generic prompt about behaving like a math tutor (without the rubric), and (4) utterances generated by Llama 3.2 3B using the same generic prompt. 
In general, method (2) contains high-quality tutor utterances that serve as positive examples that score high under our rubric. However, we found that in many cases (1) performs better for eliciting correct student responses due to the concise nature of the utterances. Utterances generated by (3) are a mix of high and low quality, while utterances generated by (4) are typically lower quality and serve as a source of negative examples for preference optimization. Prior work has shown diverse candidate quality, particularly negative examples, to improve DPO performance \cite{scarlatos2024improving,xu-etal-2024-automatic}.

After evaluating each candidate using LLMKT and the rubric, we create a combined weighted score for a candidate tutor utterance at turn $m$:
\begin{align*}
    s_m = \lambda \cdot y_m + (1 - \lambda) \cdot r_m,
\end{align*}
where $y_m$ is the probability of a correct student response predicted by LLMKT at this turn, and $r_m$ is the overall rubric score at this turn normalized in $[0,1]$. We adjust $\lambda$ to balance the tradeoff between how much the generated utterances elicit correct student responses compared to how much they follow pedagogical practice. By default, we set $\lambda=0.5$ to achieve a balance between these objectives; in our experiments, we show how varying $\lambda$ affects the balance between both.

We then use the score $s_m$ to construct preference pairs between candidate tutor utterances. We consider a candidate with score $s_m^1$ to be preferred over a candidate with score $s_m^2$ if the former is greater by some threshold, i.e., $s_m^1 - s_m^2 > \epsilon$. If the scores are within $\epsilon$, we do not form a preference pair. In practice, we set $\epsilon=0.1$ to achieve a balance where noisy preference pairs are excluded, but we retain enough data to sufficiently train the model, both of which have been shown to be important considerations for DPO \cite{kim-etal-2024-margin}.

\subsection{Model Training}
\label{sec:training}
We train our model in a two-stage process: (1) distillation and (2) DPO.
Distillation is a common way to enhance the capabilities of small LLMs by mimicking the behavior of much larger LLMs \cite{distillation}. In our case, we fine-tune Llama 3.1 8B on candidate tutor utterances generated by GPT-4o. Through this distillation stage, we gain access to a \textit{local} model that scores well on our pedagogical rubric.

We then use DPO and our preference pairs to further steer the distilled model towards effective tutoring.  
DPO trains an LLM by contrasting outputs in a preference pair given the same input prompt, using the following objective:
\begin{equation*}
    \mathcal{L}_{\text{DPO}}(\theta) = -\mathbb{E}_{(x,t^w_m,t^l_m)\sim \mathcal{D}} \left[\log\sigma\left(\beta\log\frac{\pi_\theta(t^w_m|x)}{\pi_{\text{ref}}(t^w_m|x)} - \beta\log\frac{\pi_\theta(t^l_m|x)}{\pi_{\text{ref}}(t^l_m|x)}\right)\right],
\end{equation*} 
where $t^w_m$ and $t^l_m$ represent the preferred and unpreferred tutor utterances, respectively. $x$ represents the input prompt, comprising an instruction, the context of the dialogue, and the dialogue history $t_{<m}, s_{<m}$. $\pi_\theta$ denotes the model being trained, $\pi_{\text{ref}}$ represents a frozen reference model, and $\beta$ is a hyperparameter that controls the divergence between the learned and reference policies. 

In this work, we use the distilled model as the reference model $\pi_\text{ref}$ and for initializing the weights of $\pi_\theta$. We find the distilled model works much better than using the base Llama model, since the distilled model already performs better on the rubric. Additionally, we set $\beta=0.1$, a relatively low value for DPO compared to a common value of $\beta=0.5$. The low $\beta$ value allows $\pi_\theta$ to diverge more from $\pi_\text{ref}$, which we find necessary to increase LLMKT's predictions of eliciting correct student responses in the next dialogue turn.

\section{Experimental Settings}

\subsection{Dataset}
We evaluate our framework using the MathDial dataset~\cite{macina-etal-2023-mathdial}, which consists of tutoring dialogues focused on mathematics problems from GSM8K~\cite{cobbe2021gsm8k}. Each dialogue is centered around an incorrect student solution to the math problem, and the goal of the dialogue is for the tutor to guide the student to the correct solution by addressing their misconceptions. Tutors are role-played by crowd workers, while students are simulated by GPT-3.5. Despite only being half-authentic, MathDial is the largest publicly available one-on-one tutor-student math dialogue dataset to the best of our knowledge. To estimate correctness with LLMKT, which requires knowledge component labels at the current turn, we use the annotated knowledge components from the Dialogue KT version of the dataset \cite{scarlatos2024exploring} and filter out unlabeled turns. 

We follow the original MathDial train/test split. After filtering, our test set has 588 dialogues with 3,101 tutor turns. We split the train set into a 80/20 train/validation split at the dialogue-level, resulting in 1,809/453 dialogues with 11,058/2,811 tutor turns. When creating our overgenerated tutor turn dataset for distillation and DPO training, we take a subset of the train/validation split to reduce labeling costs, resulting in 483/135 dialogues with 3,080/920 GPT-4o-generated tutor utterances for distillation and 9,662/3,095 preference pairs, with our default parameters of $\lambda=0.5$ and $\epsilon=0.1$, for DPO.

\subsection{Baselines}

We compare our preference-optimized LLM, which we refer to as \textbf{DPO}, with Llama 3.1 8B as the base model, against the following baselines: the \textbf{Base model} of Llama 3.1 8B prompted with our evaluation criteria (rubric and intending to elicit correct student responses); its supervised fine-tuned version on human tutor utterances in the original dataset, \textbf{SFT}; a fine-tuned version distilled from GPT-4o-generated utterances detailed above in Section~\ref{sec:training}, \textbf{Distill}; \textbf{GPT-4o}, the large, proprietary LLM, which is prompted with our evaluation criteria; and finally, the \textbf{Human Tutor} utterances from the original MathDial dataset.

\subsection{Automated Metrics}

We evaluate tutor utterances on both \textbf{student outcomes} and \textbf{pedagogical principles}. Student outcome prediction uses LLMKT to estimate the probability of a correct student response, averaged across turns. Evaluating how well the utterance follows the pedagogical principles will be reported in the same order as in Table~\ref{tab:ai_tutor_criteria}, reporting scores for \textbf{Acc.} (Accuracy), \textbf{Prog.} (Progress), \textbf{Err.} (Error Identification), \textbf{Hint} (Strategic Hinting), \textbf{Wth.} (Withholding), and \textbf{Enc.} (Encouraging), along with an \textbf{Overall} score.

\subsection{Model Parameters}

We use the \texttt{meta-llama/Meta-Llama-3.1-8B-Instruct} model from Hugging Face~\cite{huggingface} for all local experiments. To adapt the model, we use Low-Rank Adaptation (LoRA)~\cite{hu2021lora} with a rank parameter of \( r=64 \), scaling factor \( \alpha=32 \), and a dropout rate of 0.05. We train using the AdamW optimizer, with a learning rate of \(1 \times 10^{-4}\) for distillation and \(3 \times 10^{-5}\) for DPO, with a linear warmup phase for the first $10\%$ of training steps. We use an effective batch size of $64$ with gradient accumulation, set weight decay to \(1 \times 10^{-2}\), and set a gradient clipping maximum norm of $1.0$. We evaluate the loss on the validation set after each epoch, and achieve the minimum validation loss after three epochs for distillation and after one epoch for DPO. At test time, we generate tutor utterances using greedy decoding. We conduct all experiments on NVIDIA RTX A6000 GPUs. 

\section{Experimental Results}

\subsection{Quantitative Results}

\begin{table}[]
    \centering
    \caption{Tutor utterance evaluations for each method on the test set. The best result for each metric is in \textbf{bold} while the second best is \underline{underlined}. DPO significantly outperforms all methods on improving student correctness predictions, and closely matches the performance of GPT-4o on the pedagogical rubric.}
    \begin{tabular}{l c ccccccc}
        \toprule
         & \textbf{Student Outcomes} & \multicolumn{7}{c}{\textbf{Pedagogical Principles}}\\
        \cmidrule(lr){3-9}
        \textbf{Method} & & \textbf{Acc.} & \textbf{Prog.} & \textbf{Err.} & \textbf{Hint} & \textbf{Wth.} & \textbf{Enc.} & \textbf{Overall} \\
        \midrule
        Human Tutor & 0.45 & \textbf{0.99} & 0.66 & 0.58 & 0.49 & \underline{0.97} & 0.53 & 6.97 \\
        \rowcolor{gray!21} \multicolumn{9}{c}{GPT-4o - Proprietary LLM}\\
        Base Model & \underline{0.49} & \textbf{0.99} & \textbf{0.97} & \textbf{0.92} & \textbf{0.96} & \textbf{0.99} & \underline{0.89} & \textbf{9.40} \\
        \rowcolor{gray!21} \multicolumn{9}{c}{Llama 3.1 8B Instruct - Open-source LLM}\\
        Base Model & 0.43 & 0.82 & 0.69 & 0.70 & 0.64 & 0.94 & 0.62 & 7.20  \\
        SFT & 0.47 & 0.86 & 0.32 & 0.23 & 0.21 & 0.90 & 0.31 & 4.73 \\
        Distill & 0.47 & 0.95 & 0.92 & \underline{0.90} & \underline{0.91} & \textbf{0.99} & 0.82 & 8.93 \\
        DPO ($\lambda=0.5$) & \textbf{0.65} & \underline{0.97} & \underline{0.96} & \textbf{0.92} & \textbf{0.96} & \textbf{0.99} & \textbf{0.92} & \underline{9.37} \\
        \bottomrule
    \end{tabular}
    \label{tab:main_results}
\end{table}

Table~\ref{tab:main_results} shows the results of tutor utterance generation on the MathDial test set, across all methods on the automated metrics. We observe that DPO significantly outperforms all other methods on student correctness prediction, improving over the next best method, GPT-4o, by $33\%$. This result shows that DPO training is necessary to generate tutor utterances that are more likely to achieve correct student responses, which cannot be achieved simply through prompting. Additionally, DPO achieves similar scores to GPT-4o on the pedagogical rubric, with both methods scoring very high on all rubric items and the overall score. This result shows that our DPO training pipeline can effectively bring the pedagogical ability of small, open-source LLMs with only 8B parameters to the level of very large, proprietary LLMs.
Moreover, we see that DPO improves over distillation on almost all metrics, showing that DPO is necessary to achieve both objectives.

A notable observation is that human-written tutor utterances, while almost always accurate, score relatively low on most other rubric items. This result is not surprising since human tutors do not act according to our evaluation rubric; many of the human utterances are relatively short and simply ask the student to retry without giving guidance, and occasionally feature the tutor getting frustrated with the student. As a result, the SFT model, trained on human tutor utterances, does not score high on our rubric; the distilled model and even the base Llama model significantly outperform it on the rubric items.

\begin{figure}
    \centering
    \includegraphics[width=.7\linewidth]{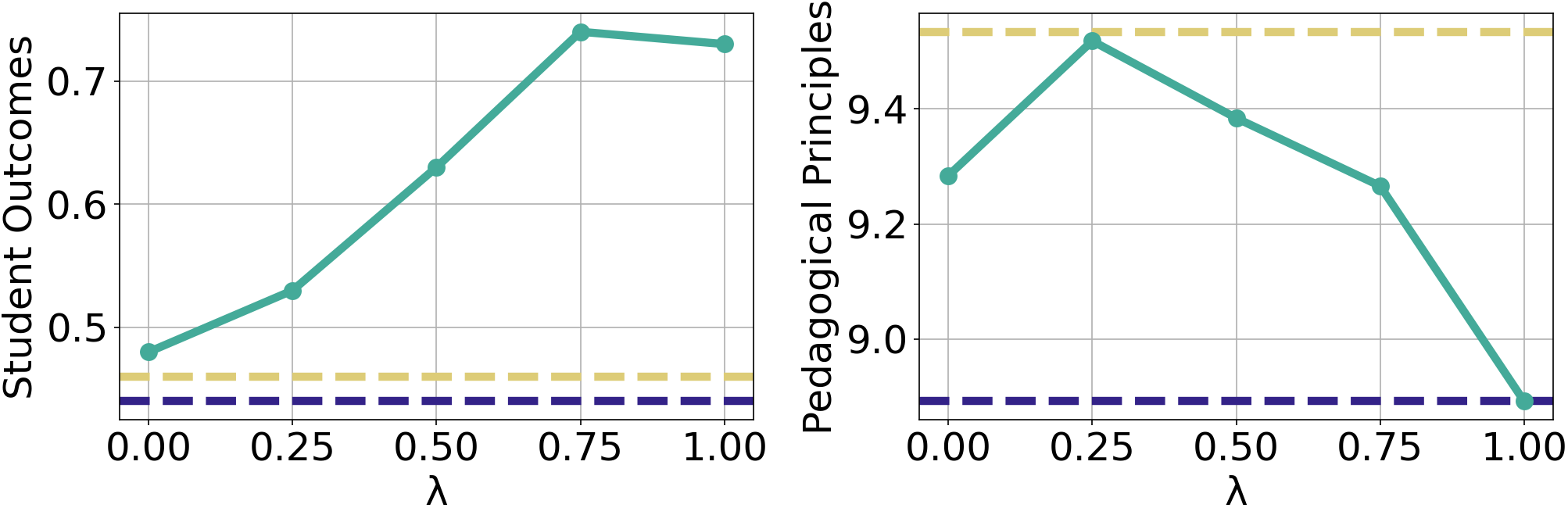}
    \caption{Results of our experiment adjusting the value of $\lambda$ in \textcolor{dpo-color}{DPO} training. Left: student correctness prediction changing with $\lambda$. Right: pedagogical rubric score changing with $\lambda$. We also show values for \textcolor{gpt-4o-color}{GPT-4o} and the \textcolor{distill-color}{Distill} model for comparison.}
    \label{fig:lambda}
\end{figure}

\noindent\textbf{Student Outcomes vs.\ Pedagogical Principles}
We investigate how adjusting the value of $\lambda$ can balance the two goals of maximizing student outcomes under LLMKT and following pedagogical principles. We vary the value of $\lambda \in \{ 0.00, 0.25, 0.50, 0.75, 1.00 \}$, where larger values attribute more weight to the student correctness prediction objective, and evaluate on a subset of the data with 500 turn pairs. Figure~\ref{fig:lambda} shows the result of this experiment. We see that, as expected, increasing $\lambda$ from $0$ to $1$ generally increases student correctness prediction performance while generally decreasing the rubric score. Practitioners can get their desired balance by changing $\lambda$; for example, a value of around $0.5$ to $0.75$ can maintain a relatively high score on the pedagogical rubric while significantly improving performance on soliciting correct responses from students. Perhaps surprisingly, changing $\lambda$ from $0.75$ to $1.0$ does not increase student outcome performance, while decreasing $\lambda$ from $0.25$ to $0.0$ \textit{decreases} pedagogical performance. This result implies that a small reward signal from correctness prediction may implicitly help pedagogical effectiveness, and vice versa.

\subsection{Qualitative Analysis}
\label{sec:qual-eval}

We conduct a qualitative analysis of generated tutor utterances to investigate the strengths and weaknesses of our method, how it compares to baselines, and what types of patterns emerge in the model-generated text.

Table~\ref{tab:qual_analysis} shows an example dialogue context and tutor utterances from the human tutor, GPT-4o, and DPO.
We see that the human tutor states that the student is not correct and prompts the student to try again, providing a subtle hint to focus on the cost of the dozen. However, this utterance neither gives LLMKT, the student model, any confidence in the student being able to answer correctly, nor follows strong pedagogical principles. GPT-4o guides the student toward correctly calculating the profit, but directly provides half of the solution in its hint, limiting deep student engagement. It also poses the challenging task of computing the total profit, which requires multiple calculation steps. Since the student is still struggling with unit conversion, LLMKT is not confident that the student will respond correctly. In contrast, our DPO-trained model provides an actionable hint for the student, following the rubric. It also poses the simpler task of asking the student to first find the cost of a half dozen, which is necessary for later steps. As a result, LLMKT is confident the student can respond correctly.

Overall, we find that our DPO-trained model excels at posing \textit{concrete} tasks that are nontrivial, but more feasible for students compared to GPT-4o. We observe that the behavior of posing tasks this way is the primary way of achieving higher predictions for student correctness, and appears to be an emergent behavior of training on the LLMKT-derived scores. While the tasks sometimes only ask the student to perform simple arithmetic, they are typically more involved. For example, in another dialogue, the model asks ``How long does it take to download 200 GB at a rate of 2 GB/minute?'' This task requires the student to set up a single variable equation and understand the roles of the given values. We also find that the DPO-trained model is more likely to ask questions, whereas other methods often directly tell the student to perform a task.

However, both LLMs still have some limitations, particularly when students give partially correct responses. In these cases, the models may ignore the progress the student has made and focus on errors that are irrelevant to the recent student response. Correctly identifying errors in math with LLMs has been shown to be challenging in prior work as well~\cite{fernandez-etal-2024-divert,liu2023novice}, indicating additional techniques may be needed to fix this behavior.

\begin{table}[h]
    \centering
    \caption{Comparing human tutor, GPT-4o, and DPO tutor utterances, on student outcomes (SO) and pedagogical principles (PP). \textcolor{brown}{Brown} highlights the tutor-posed task. Some irrelevant text in the dialogue context has been removed for brevity.
    }
    \scalebox{.7}{
    \begin{tabular}{p{1.3cm}p{18cm}@{\hskip .5cm}p{.8cm}p{.8cm}}
        \toprule
        \rowcolor{gray!21} \multicolumn{4}{c}{Problem} \\        \multicolumn{4}{p{21.4cm}}{Chloe bought chocolate-dipped strawberries at \$50 a dozen. She then sold them for \$30 for half a dozen. How much is Chloe's profit if she sold 50 dozens?} \\
        \rowcolor{gray!21} \multicolumn{4}{c}{Dialogue History} \\
        $t_1$ & \multicolumn{3}{p{20cm}}{Hey Alejandra.  If Chloe buys the strawberries at \$50 for 12 and sells them for \$30 for 6, how much profit would that be for one dozen?} \\
        $s_1$ & \multicolumn{3}{p{20cm}}{For one dozen, Chloe would make a profit of \$20, which is \$50 - \$30 = \$20.} \\
        $t_2$ & \multicolumn{3}{p{20cm}}{Remember that she's selling half a dozen for \$30.  How many half dozen can she sell from each \$50 dozen she buys?} \\   
        $s_2$ & \multicolumn{3}{p{20cm}}{She can sell two half dozen for each \$50 dozen she buys, so she can make a profit of \$60 from each \$50 dozen.} \\        
        \rowcolor{gray!21} \multicolumn{4}{c}{Next Tutor Turn ($t_3$) Candidates} \\
        \textbf{Method} & \textbf{Tutor Utterance} & \textbf{SO} & \textbf{PP} \\
        \midrule
        Human Tutor & Remember it's not \$60 profit, because they cost her to buy them. \textcolor{brown}{Try again.} & 0.33 & 4 \\
        GPT-4o & It looks like there's a misunderstanding in your calculations. Remember, Chloe's cost for half a dozen is \$25 (since \$50 per dozen divided by 2). She sells half a dozen for \$30, so her profit per half dozen is \$30 - \$25 = \$5. \textcolor{brown}{Try recalculating her total profit using this correct profit per half dozen.} & 0.59 & \textbf{10}\\
        DPO & It looks like there's a misunderstanding. If Chloe buys a dozen for \$50 and sells half a dozen for \$30, let's first find out how much she paid for half a dozen. \textcolor{brown}{If \$50 is for 12, how much is \$50 divided by 2?} & \textbf{0.84} & \textbf{10} \\
        \bottomrule
    \end{tabular}}
    \label{tab:qual_analysis}
\end{table}

\subsection{Human Evaluation}
\label{sec:human-eval}

We also conduct a human evaluation to further assess different tutor models and how our automated metrics align with human judgment. 

We recruit two independent volunteer university students to annotate tutor utterances according to our evaluation metrics.
We randomly sample 10 dialogues from the test set after filtering based on toxicity and low quality ground-truth tutor utterances.
The annotators evaluate 5 consecutive tutor turns from each dialogue, resulting in a total of 50 evaluation instances. At each instance, we show participants the dialogue so far and next tutor utterance candidates generated by (1) human tutors, (2) GPT-4o, and (3) our DPO-trained model, in random order without revealing the method. 

For student outcome prediction, annotators rank the three candidate utterances based on how likely they think each one will lead to a correct student response. We ask annotators to take into account both the task posed in the turn and the student's knowledge demonstrated in prior turns. We use rankings instead of absolute values since the latter may be harder for humans to calibrate. 
For pedagogical principles, we simply ask annotators to follow the same evaluation rubric outlined in Table~\ref{tab:ai_tutor_criteria} and provide an overall score on a 1-10 scale.

\begin{table}[h]
    \centering
    \caption{Human evaluation shows that DPO outperforms human tutors and GPT-4o.}
    \begin{tabular}{l c c}
        \toprule
        \textbf{Method} & \textbf{Correctness Rank}$^\downarrow$ & \textbf{Rubric Score}$^\uparrow$ \\
        \midrule
        Human Tutor & 2.12 & 7.36 \\
        GPT-4o & 2.13 & 8.07 \\
        DPO & \textbf{1.75} & \textbf{8.55} \\
        \bottomrule
    \end{tabular}
    \label{tab:human-scores}
\end{table}

\noindent\textbf{Results} Table~\ref{tab:human-scores} shows average student correctness ranks (rank $1$ means top-ranked) and rubric scores from human annotators. We see that DPO outperforms both human tutors and GPT-4o on both metrics, with all differences statistically significant ($p<0.05$) according to a paired t-test. Higher correctness ranks show that DPO learns how to pose tasks that are manageable to students, while the higher rubric score shows that it does so while maintaining pedagogical principles. Additionally, the higher rubric score may imply that the correctness objective also indirectly improves pedagogical quality.
While DPO outperforming GPT-4o on the rubric is contrary to the automated results in Table~\ref{tab:main_results}, the automated results may reflect self-bias from GPT-4o as the evaluator \cite{xu-etal-2024-pride}.

Since these evaluation tasks can be highly subjective, we also investigate the inter-rater agreement between the two human annotators and the automated metrics. We use Kendall's $\tau$~\cite{kendall1948rank} for correctness ranks and Pearson's correlation coefficient $\rho$ ~\cite{lee1988thirteen} for rubric scores. For correctness ranks, Kendall's $\tau$ is only $0.06$ between the two annotators and averages $0.08$ between LLMKT and both annotators. This low agreement is not surprising due to the subjective nature of the task; we find that there are many near-ties between different candidate utterances according to LLMKT's output probabilities. For rubric scores, the inter-rater agreement is notably higher: Pearson's $\rho$ is $0.15$ between the two annotators and averages $0.27$ between GPT-4o and both annotators, the latter being statistically significant. This result suggests that evaluating pedagogical principles is much easier than predicting student correctness, though the former is still subjective, since annotators may value certain rubric items differently. 

Overall, despite showing some promise through preliminary evaluation, it is important to deploy our trained tutor LLMs to real students and test whether they actually lead to good student learning outcomes.

\section{Conclusions and Future Work}

In this paper, we introduced a methodology to train LLM tutors to maximize student outcomes in dialogues while maintaining high pedagogical quality. We use student simulation to predict how likely a tutor utterance will yield a correct student response, and use GPT-4o to evaluate the pedagogical quality of tutor utterances using a rubric. We use both sets of predictions to score overgenerated tutor utterances and fine-tune Llama 3.1 8B with direct preference optimization. Our resulting model significantly outperforms other methods for increasing predicted student outcomes, and matches the performance of GPT-4o, a much larger closed model, on pedagogical aspects. There are many avenues for future work. First, an evaluation with real students should be carried out to determine if our methods are still effective in real-world scenarios. Second, future work should investigate how to optimize for longer-term learning outcomes, such as concept mastery or performance on post-dialogue assessments. Third, future work should include estimates of student affect and engagement as part of the reward. Finally, future work should apply our method to dialogues in non-math domains, such as language learning or computer science.

\section*{Acknowledgments}

We thank Adriana Caraeni and Henry Yang for helpful discussions around this work. This work is partially supported by Renaissance Philanthropy via the learning engineering virtual institute (LEVI) and NSF grants 2118706, 2237676, and 2341948.

\bibliographystyle{plain}
\bibliography{main}

\end{document}